\documentclass{article}
\usepackage{spconf,amsmath,graphicx,hyperref}

\usepackage{booktabs}
\usepackage[table]{xcolor}
\usepackage{hyperref}       
\hypersetup{
    colorlinks,
    linkcolor={red!90!black},
    citecolor={green!90!black},
    urlcolor={blue!30!black}
}
\usepackage{url}            
\usepackage{amssymb}

\usepackage[tracking=true]{microtype} 
\SetTracking{encoding=*}{-20}
\AtBeginEnvironment{thebibliography}{
  \begingroup
  \microtypesetup{tracking=true}
  \lsstyle
}
\AtEndEnvironment{thebibliography}{\endgroup}

\AtBeginEnvironment{thebibliography}{%
  \linespread{0.86}\selectfont
}

\title{Robust Grounding with MLLMs against Occlusion and Small Objects \\
via Language-Guided Semantic Cues}
\name{Beomchan Park$^\star$, Seongho Kim$^\star$, Hyunjun Kim, Sungjune Park, Yong Man Ro$^\dagger$\thanks{$^\star$Equal Contribution. $^\dagger$Corresponding Author.}}
\address{
Integrated Vision Language Lab., KAIST, Republic of Korea\\
\small{\texttt{\{bpark0810, taho43, kimhj709, sungjune-p, ymro\}@kaist.ac.kr}}
}
\begin{document}

\maketitle

\begin{abstract}
\looseness=-1
While Multimodal Large Language Models (MLLMs) have enhanced grounding capabilities in general scenes, their robustness in crowded scenes remains underexplored. Crowded scenes entail visual challenges (\textit{i.e.,} occlusion and small objects), which impair object semantics and degrade grounding performance. In contrast, language expressions are immune to such degradation and preserve object semantics. In light of these observations, we propose a novel method that overcomes such constraints by leveraging Language-Guided Semantic Cues (LGSCs). Specifically, our approach introduces a Semantic Cue Extractor (SCE) to derive semantic cues of objects from the visual pipeline of an MLLM. We then guide these cues using corresponding text embeddings to produce LGSCs as linguistic semantic priors. Subsequently, they are reintegrated into the original visual pipeline to refine object semantics. Extensive experiments and analyses demonstrate that incorporating LGSCs into an MLLM effectively improves grounding accuracy in crowded scenes.
\end{abstract}


\vspace{-1.0mm}
\section{Introduction}
\label{sec:intro}
\vspace{-1.0mm}

\begin{figure*}[t]
    \centering
    \includegraphics[width=\textwidth]{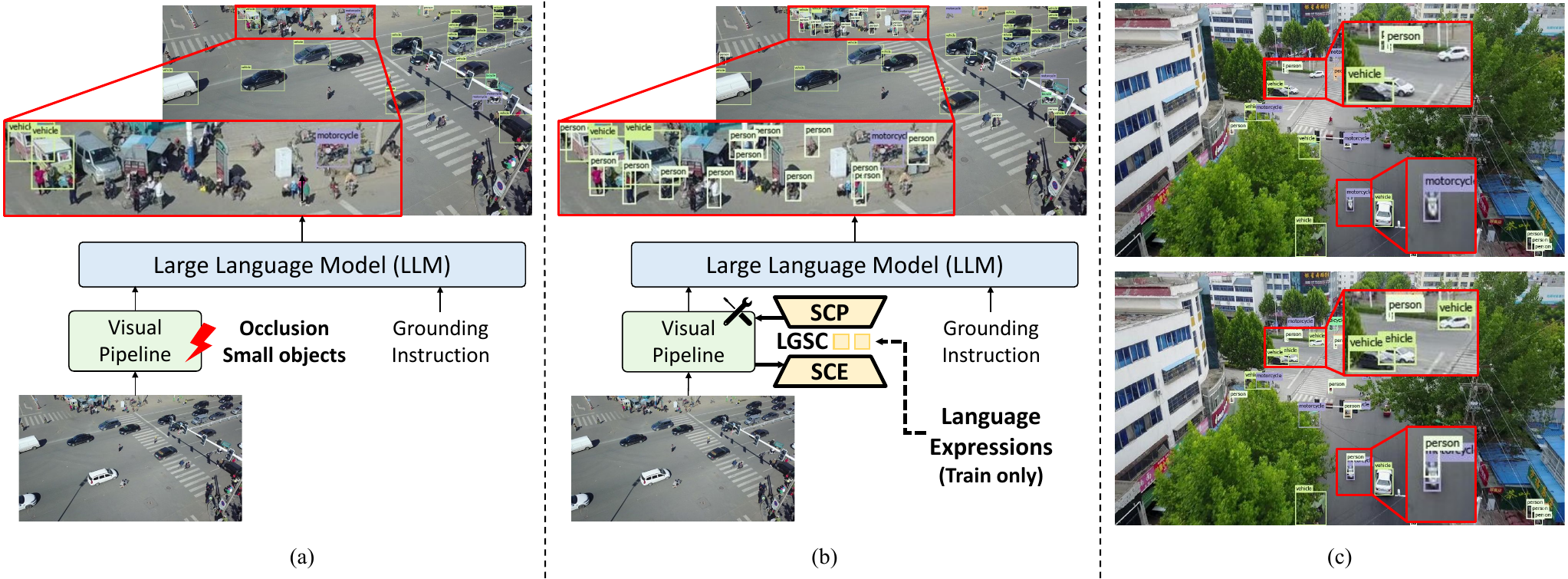}
    \vspace{-0.8cm}
    \caption{(a) An existing grounding MLLM~\cite{jiang2024chatrex} shows degraded performance in crowded scenes due to occlusion and small objects. (b) In contrast, our approach trains LGSCs to handle these constraints and successfully captures crowded objects. (c) Additional qualitative results demonstrate that our approach excels in crowded scenes.}
    \label{fig:motivation}
    \vspace{-0.2cm}
\end{figure*}

Multimodal Large Language Models (MLLMs)~\cite{bai2025qwen2, zhu2025internvl3} have recently achieved remarkable progress in diverse multimodal tasks by interactively following human instructions. Central to this success is visual grounding~\cite{wang2023visionllm, peng2024grounding, you2023ferret}, the ability to connect visual objects with referring language expressions, such as object categories or captions. While prior studies~\cite{ma2024groma, jiang2024chatrex} have primarily focused on enhancing the grounding capabilities of MLLMs in general scenes~\cite{yu2016modeling, mao2016generation}, the challenge of maintaining robustness in crowded scenes remains largely underexplored. Crowded scenes are frequently encountered in critical real-world scenarios, including public safety surveillance and traffic monitoring~\cite{shao2018crowdhuman, du2019visdrone, du2018unmanned, sun2025refdrone}. Yet, recent state-of-the-art models struggle to accurately ground language expressions to visual objects in crowded scenes, as shown in Fig.~\ref{fig:motivation}. Crowded scenes are characterized by occlusion and small objects that create severe visual challenges. These factors distort visual information, compromising object semantics and substantially degrading the grounding performance of existing MLLMs.

On the other hand, language expressions are naturally immune to such visual constraints and preserve object semantics. In light of these observations, we propose a novel method that enhances MLLM robustness in crowded scenes by leveraging Language-Guided Semantic Cues (LGSCs). Our approach exploits language annotations of visual objects in the training data as a corrective source of semantic supervision for degraded visual information. This idea is motivated by neuroscience research~\cite{lupyan2013language, boutonnet2015words} suggesting that language labels provide a top-down boost to visual awareness under adverse conditions by pre-activating or imposing categorical priors on visual processing in the human brain. Aligned with these findings, prior work in machine learning~\cite{kim2024weather, park2025language} further demonstrates that incorporating language semantics can refine visual representations under such constraints. Building on these insights, we introduce an auxiliary branch into the MLLM visual pipeline to extract semantic cues from objects and refine them with language expressions to produce LGSCs.

Concretely, we first extract text embeddings of ground-truth language expressions using a pretrained text encoder. We then employ a Semantic Cue Extractor (SCE) to derive semantic cues from object features. Mirroring the mechanism of human cognition, we guide these cues using the corresponding text embeddings to internalize linguistic semantic priors, thereby forming LGSCs. Subsequently, they are reintegrated into the visual pipeline via a Semantic Cue Projector (SCP) and a low-rank cross-attention block~\cite{lee2024moai}. At inference, these LGSCs are generated without external language input, acting as learned priors to refine degraded object semantics. This simple yet effective process significantly enhances the grounding robustness of MLLMs in crowded scenes.

\looseness=-1
We evaluate our method under both supervised and zero-shot settings on diverse, widely used benchmarks specializing in crowded scenes~\cite{shao2018crowdhuman, du2019visdrone, du2018unmanned, sun2025refdrone, zhang2017citypersons, zhang2019widerperson, feng2024hazydet}. Extensive experiments and analyses demonstrate that our approach consistently improves grounding accuracy across all datasets. In summary, our contributions are as follows: \textbf{i)} Drawing on insights from neuroscience, we propose the SCE and SCP to extract, guide, and reintegrate LGSCs. This mechanism boosts the recognition of occluded and small objects with linguistic semantic priors, enhancing the grounding robustness of MLLMs in crowded scenes. \textbf{ii)} Through comprehensive experiments on challenging benchmarks, we validate that our approach effectively grounds various visual objects in crowded scenes, demonstrating superior performance across diverse settings.

\vspace{-1mm}
\section{Methodology}
\label{sec:method}
\vspace{-1.0mm}

\subsection{Overall Architecture}
\label{subsec:overall-arch}
\begin{figure*}[!t]
    \centering
    \includegraphics[scale=0.56]{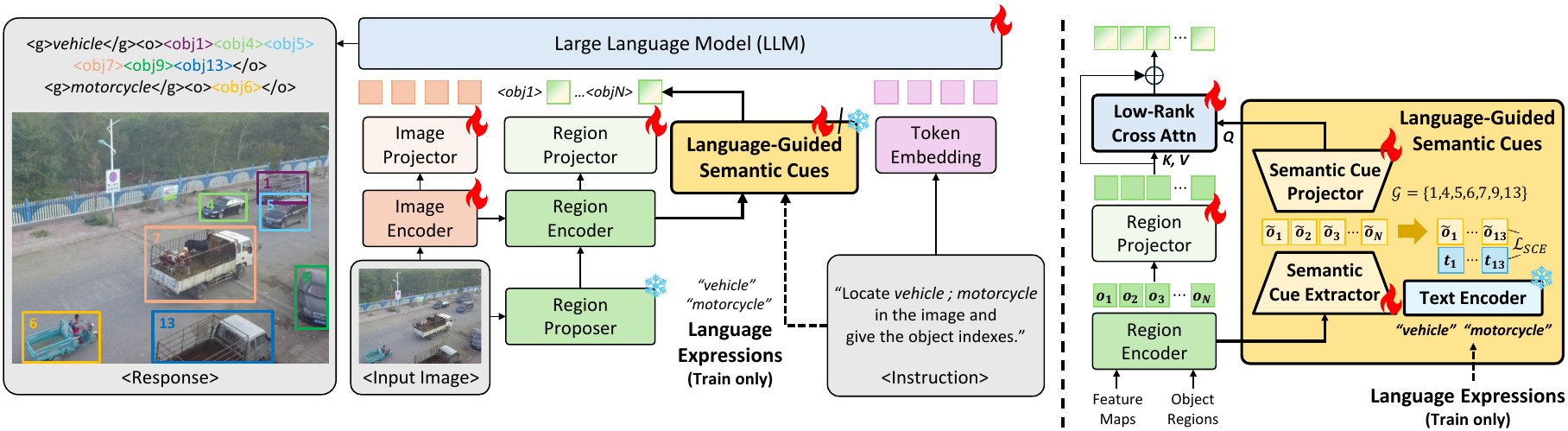}
    \vspace{-0.1cm}
    \caption{Overall architecture of our approach (left) and detailed illustration of training Language-Guided Semantic Cues (right).}
    \label{fig:method}
    \vspace{-0.1cm}
\end{figure*}

The left side of Fig.~\ref{fig:method} illustrates the overall architecture of our approach. We adopt ChatRex-7B~\cite{jiang2024chatrex} as our baseline, which follows the decoupled localization paradigm~\cite{ma2024groma} for effective grounding with MLLMs. Specifically, the architecture utilizes an image encoder~\cite{dosovitskiy2020image, liu2022convnet} to extract image feature maps. These maps are flattened and passed through an image projector to produce image tokens, which are then fed into the LLM~\cite{chiang2023vicuna} alongside tokenized instructions.

Furthermore, the model incorporates a region proposer~\cite{jiang2024t, zhao2024detrs} to predict object regions from the input image. It is a class-agnostic object detector trained on a single category, and filters predicted regions based on non-maximum suppression and objectness scores when their number exceeds the predefined maximum $N$. A region encoder then extracts object features by performing RoIAlign~\cite{he2017mask} on the image feature maps corresponding to these regions. These object features are processed by a region projector to generate object tokens, which are indexed by special tokens (\textit{e.g.,} \texttt{<obj1>,...,<objN>}) before being fed to the LLM. Consequently, given a grounding instruction containing referring language expressions, the LLM identifies the target objects via index prediction. 

Building upon this baseline, we introduce Language-Guided Semantic Cues (LGSCs) into the visual pipeline. Inspired by linguistic top-down semantic priors in human cognition, our method learns to extract semantic cues whose high-level semantics are guided by text embeddings of language expressions during training. In the following section, we elaborate on how LGSCs are generated and trained. 

\subsection{Semantic Cue Extraction and LGSCs}
To extract semantic cues and guide them using language expressions to generate LGSCs, we introduce the Semantic Cue Extractor (SCE), implemented as a 4-layer MLP with GELU activation~\cite{hendrycks2016gaussian}. This module maps object features $\boldsymbol{o}_i \in \mathbb{R}^{d_o}$ from the visual pipeline to semantic cues $\boldsymbol{\tilde{o}}_i \in \mathbb{R}^{d_t}$, where $d_o$ and $d_t$ are the dimensions of the object features and semantic cues, respectively. To align these cues with linguistic semantic priors, we utilize a pretrained text encoder~\cite{laion_CLIP_convnext_large_d_320_laion2B_s29B_b131K_ft_soup} to generate text embeddings from ground-truth language expressions (\textit{e.g.,} object categories or captions), which serve as a corrective source of semantic supervision.

\looseness=-1
During training, we combine object regions predicted by the region proposer with ground-truth object regions into a unified set, as illustrated on the right side of Fig.~\ref{fig:method}. Object features extracted from these combined regions are randomly shuffled to ensure the model's robustness against positional biases. Let $\mathcal{G}$ denote the set of indices corresponding to ground-truth object regions. For each semantic cue $\boldsymbol{\tilde{o}}_i$ associated with an index $i \in \mathcal{G}$, we obtain the corresponding text embedding $\boldsymbol{t}_i \in \mathbb{R}^{d_t}$ from the text encoder and maximize the cosine similarity between them. To this end, we define the SCE loss as:
\begin{equation}
    \mathcal{L}_{\mathrm{SCE}}=\frac{1}{|\mathcal{G}|}\sum_{i\:\in\:\mathcal{G}}\!\left(1-\frac{\boldsymbol{\tilde{o}}_i^{\top} \ \boldsymbol{t}_i}{\lVert\boldsymbol{\tilde{o}}_i\rVert_2\,\lVert\boldsymbol{t}_i\rVert_2}\right).
\end{equation}

This loss encourages the SCE to map object features into LGSCs that internalize linguistic semantic priors, even in the absence of language expressions at inference time. These cues are subsequently reintegrated into the original visual pipeline to refine object semantics degraded by occlusion and small objects in crowded scenes, as detailed in the next section.

\subsection{Object Semantic Refinement with LGSCs}
\looseness=-1
Once LGSCs are extracted and optimized via $\mathcal{L_\text{SCE}}$, they are passed through a Semantic Cue Projector (SCP), implemented as a 2-layer MLP with GELU activation, to map them into the input embedding space of the backbone LLM. In parallel, the original object features are also projected into the same embedding space via a region projector. Given that LGSCs are aligned with text embeddings, this projection enables effective interaction within the LLM's embedding space, which is inherently structured around language semantics.

To refine object semantics degraded by occlusion and small size, we integrate these cues using a low-rank cross-attention block~\cite{vaswani2017attention,lee2024moai} equipped with an external residual path. In this mechanism, the projected LGSCs serve as queries, while the projected object features act as keys and values. This configuration allows the LGSCs to selectively attend to and aggregate relevant visual contexts from the original object features. To minimize computational overhead while preserving expressive interactions, we employ a low-rank multi-head cross-attention mechanism, enabling the model to effectively enhance representation quality with minimal additional cost.

\subsection{Objective Functions}
Finally, the LLM accepts the image features $\mathcal{I}$, the refined object features $\mathcal{O}$, and the tokenized grounding instruction $\mathcal{T}$ as input. The LLM processes this multimodal input $\mathcal{M} = (\mathcal{I}, \mathcal{O}, \mathcal{T})$ to generate the grounding output $y$ in the following format: \texttt{<g>phrase</g><o><objK>...<objL></o>}. The model is trained using the following autoregressive loss:
\begin{equation}
    \mathcal{L}_{\mathrm{AR}}=-\sum^{T}_{t=1}\log p\!\left(y_t \,\middle|\, \mathcal{M}, y_{<t}\right).
\end{equation}

The final training objective combines this autoregressive loss with the SCE loss: $\mathcal{L} = \mathcal{L}_{\mathrm{AR}} + \lambda \mathcal{L}_{\mathrm{SCE}}$, where $\lambda$ is a hyperparameter balancing the two objectives. This joint optimization enables the model to accurately generate grounding outputs while enriching object representations in crowded scenes with LGSCs.

\vspace{-0.1cm}
\section{Experiments}
\label{sec:experiments}
\vspace{-0.1cm}

\subsection{Datasets}
We first train and evaluate our method under a supervised setting on four widely used benchmarks specializing in crowded scenes: CrowdHuman~\cite{shao2018crowdhuman}, VisDrone~\cite{du2019visdrone}, UAVDT~\cite{du2018unmanned}, and RefDrone~\cite{sun2025refdrone}. CrowdHuman is a human detection benchmark featuring an average of 23 persons per image, comprising 15,000 training and 4,370 validation images. VisDrone is a drone-based detection dataset containing 6,471 training and 548 validation aerial images of busy urban scenes. Although it originally includes 10 categories, we consolidate them into five (\textit{person, people, vehicle, bicycle, and motorcycle}) to mitigate category redundancy. UAVDT is another drone-based dataset consisting of 24,143 training and 16,592 validation images. Its three original categories (\textit{car, truck, bus}) are merged into a single \textit{vehicle} category to address severe class imbalance that could distort evaluation results. Notably, this class regrouping strategy is applied uniformly across all compared models to ensure a fair comparison, without conferring any specific advantage to our method. RefDrone is a Referring Expression Comprehension (REC)~\cite{mao2016generation} benchmark derived from VisDrone, maintaining the same data split. It provides descriptive captions, such as ``The pedestrians carrying umbrellas'' or ``The red vehicles in the vicinity of the gas station.'' For evaluation, we adopt the widely used $\textbf{AP}_{50}$~\cite{lin2014microsoft} metric to measure grounding performance.

\subsection{Implementation Details}
\looseness=-1
In addition to comparisons with recent grounding-capable MLLMs, we evaluate two training setups: (i) a baseline fine-tuned naively only with $\mathcal{L}_{\mathrm{AR}}$~\cite{jiang2024chatrex}, and (ii) our proposed method, which incorporates $\mathcal{L}_{\mathrm{SCE}}$ into the training objective. To robustly capture object regions across various scales, we supplement the existing region proposer~\cite{jiang2024t} with an additional module based on RT-DETR~\cite{zhao2024detrs}, fine-tuned on CrowdHuman, VisDrone, and UAVDT. To obtain text embeddings for LGSCs, we utilize the text encoder from a CLIP-pretrained ConvNeXt model trained on LAION-2B~\cite{laion_CLIP_convnext_large_d_320_laion2B_s29B_b131K_ft_soup}.

During training, we freeze the region proposer and text encoder while fine-tuning all other modules. We train for 1 epoch with a warmup ratio of 0.03 and a global batch size of 128, utilizing 8 NVIDIA A6000 GPUs. Optimization is performed using AdamW~\cite{loshchilov2017decoupled} with $\beta_1=0.9$ and $\beta_2=0.999$. We set the learning rate to 1e-4 for the SCE, SCP, and low-rank cross-attention block, while assigning 2e-5 to the remaining modules. The low-rank cross-attention mechanism is configured with a rank of 512 and 8 attention heads. Finally, the loss-balancing hyperparameter is set to $\lambda=2.0$.

\begin{table}[t!]
    \centering
    \resizebox{1\linewidth}{!}{%
    \begin{tabular}{@{}lcccc@{}}
        \toprule
        \textbf{Model} & \textbf{CH (val)} & \textbf{VD (val)} & \textbf{UAVDT (test)} & \textbf{RD (val)} \\ \midrule
        Ferret-7B~\cite{you2023ferret} & 7.2 & 0.5 & 1.9 & 11.5 \\
        InternVL3-8B~\cite{zhu2025internvl3} & 9.9 & 0.3 & 1.5 & 8.3 \\
        Qwen2.5-VL-7B~\cite{bai2025qwen2} & 15.3 & 0.8 & 7.9 & 43.2 \\
        Groma-7B~\cite{ma2024groma} & 32.1 & 6.9 & 18.4 & 38.9 \\
        ChatRex-7B~\cite{jiang2024chatrex} (baseline) & 52.0 & 19.9 & 58.2 & 39.8 \\
        \rowcolor[HTML]{EFEFEF} 
        +Naive finetuning & 67.4 & 22.7 & 67.1 & 51.5 \\
        \rowcolor[HTML]{EFEFEF} 
        +Proposed method & \textbf{72.8 (+5.4)} & \textbf{24.7 (+2.0)} & \textbf{69.9 (+2.8)} & \textbf{52.8 (+1.3)} \\ \bottomrule
        \end{tabular}
    }
    \caption{
    Grounding performance comparisons on CrowdHuman (CH), VisDrone (VD), UAVDT, and RefDrone (RD). For UAVDT, we sample 500 images from the test split for simplicity, as it contains successive redundant video frames.
    }
    \label{tab:tab1}
\end{table}

\begin{table}[t!]
    \centering
    \resizebox{0.9\linewidth}{!}{%
        \begin{tabular}{@{}lccc@{}}
        \toprule
        \textbf{Model} & \textbf{CityPersons} & \textbf{WiderPerson} & \textbf{HazyDet} \\ \midrule
        ChatRex-7B~\cite{jiang2024chatrex} & 19.5 & 50.2 & 38.3 \\
        \rowcolor[HTML]{EFEFEF} 
        +Proposed Method & \textbf{32.1 (+12.6)} & \textbf{55.8 (+5.6)} & \textbf{59.3 (+21.0)} \\
        \bottomrule
        \end{tabular}
    }
    \caption{Zero-shot grounding performance comparisons on CityPersons, WiderPerson, and HazyDet.}
    \label{tab:tab2}
    \vspace{-0.4cm}
\end{table}

\subsection{Experimental Results}
\noindent \textbf{Grounding Performance.} To highlight the challenges of crowded scenes and validate our approach, we compare it against recent state-of-the-art grounding MLLMs~\cite{you2023ferret, ma2024groma, bai2025qwen2, zhu2025internvl3} and a naively fine-tuned baseline. Tab.~\ref{tab:tab1} reveals that recent MLLMs struggle in crowded scenes despite being trained on massive grounding datasets. This observation confirms our premise that grounding with MLLMs in crowded scenes remains largely underexplored. Notably, our method consistently outperforms both recent MLLMs and the baseline across all benchmarks. Qualitative results in Fig.~\ref{fig:motivation} further illustrate these improvements. Overall, the results verify that leveraging LGSCs effectively enhances robustness against occlusion and small objects in diverse crowded environments.

\looseness=-1
\textbf{Zero-shot Performance.} Extending our evaluation beyond the supervised setting, we further assess the grounding performance of our approach in the zero-shot setting. To this end, we utilize another set of three widely used benchmarks for crowded scenes: CityPersons~\cite{zhang2017citypersons}, WiderPerson~\cite{zhang2019widerperson}, and HazyDet~\cite{feng2024hazydet}. CityPersons is a pedestrian-detection dataset featuring self-driving scenarios in urban areas, whereas WiderPerson encompasses a wide range of pedestrian situations and poses, such as marathons, traffic, and dance. These datasets thus provide crowded scenes distinct from CrowdHuman. Additionally, HazyDet is a drone-view dataset capturing hazy environments, introducing a new challenge compared to VisDrone and UAVDT. Despite not being trained on these three datasets, our proposed method substantially outperforms the baseline model in the zero-shot setting, as seen in Tab.~\ref{tab:tab2}. These results confirm that our approach can effectively generalize to handle occlusion and small objects across various crowded scenes.

\begin{table}[t!]
\centering
\resizebox{0.9\linewidth}{!}{%
\begin{tabular}{@{}lcccc@{}}
\toprule
\textbf{Model} & \multicolumn{2}{c}{\textbf{RefDrone}} & \multicolumn{2}{c}{\textbf{RefCOCOg}} \\ \midrule
 & \textbf{METEOR} & \textbf{CIDEr} & \textbf{METEOR} & \textbf{CIDEr} \\ \midrule
Groma-7B~\cite{ma2024groma} &  12.3 & 25.5 & \textbf{16.7} & 108.1\\
ChatRex-7B~\cite{jiang2024chatrex} & 7.5 & 8.9 & 13.6 & 56.8 \\
\rowcolor[HTML]{EFEFEF} 
            +Naive finetuning & 25.3 & 140.8 & 16.2 & 110.7 \\
\rowcolor[HTML]{EFEFEF} 
            +Proposed method & \textbf{26.0 (+0.7)} & \textbf{161.9 (+21.1)} & 16.2 (+0.0) & \textbf{114.0 (+3.3)} \\
\bottomrule
\end{tabular}
}
\caption{Region captioning performance comparisons on RefCOCOg and RefDrone.}
\vspace{-0.5cm}
\label{tab:tab3}
\end{table}

\textbf{Region Captioning Performance.} To fully leverage the generative capabilities of the backbone LLM, we extend our analysis to region captioning. In this task, the model is required to generate a caption for a localized visual object, where the ability to accurately align visual objects with language expressions is also critical. We employ two Referring Expression Generation (REG) benchmarks: RefDrone and RefCOCOg~\cite{yu2016modeling}. Specifically, we include RefCOCOg to show that our method is also effective for general scenes, complementing our analysis on crowded scenes. To evaluate performance, we utilize the METEOR~\cite{banerjee2005meteor} and CIDEr~\cite{vedantam2015cider} metrics, which are commonly used to comprehensively assess generated captions. As shown in Tab.~\ref{tab:tab3}, our method outperforms previous MLLMs and the fine-tuned baseline, demonstrating significant gains on RefDrone and consistent improvements on RefCOCOg. This confirms that our approach effectively enhances grounding robustness of MLLMs in understanding as well as localization.

\section{Conclusion}
Recognizing that language expressions are naturally immune to visual degradation and preserve object semantics, we propose a novel neuroscience-inspired framework leveraging Language-Guided Semantic Cues (LGSCs) via the Semantic Cue Extractor (SCE) and Semantic Cue Projector (SCP). These modules derive and integrate lingusitic semantic priors, enhancing grounding robustness against occlusion and small objects in crowded scenes. Extensive experiments on crowded-scene benchmarks, under both supervised and zero-shot settings, demonstrate significant improvements in grounding accuracy. These findings underscore the effectiveness of linguistic priors for robust multimodal understanding in complex environments.

\vfill

\pagebreak
\clearpage

\section{Acknowledgments}
\vspace{-0.2cm}
This work was partially supported by Center for Applied Research in Artificial Intelligence (CARAI) and  Hanwha Aerospace. Additionally, the supercomputing resource was partially supported by KSC (KSC-2025-CRE-0090).

\section{Compliance with Ethical Standards}
\vspace{-0.2cm}
This research was conducted retrospectively using only publicly available datasets, without private or identifiable data.

\bibliographystyle{IEEEbib}
\bibliography{refs}

\end{document}